# Utilizing Satellite Imagery Datasets and Machine Learning Data Models to Evaluate Infrastructure Change in Undeveloped Regions

| | |
|---|---|
| **Kyle McCullough, Andrew Feng, Meida Chen** | **Ryan McAlinden** |
| **University of Southern California Institute for Creative Technologies** | **Army Futures Command** |
| Playa Vista, California | Orlando, Florida |
| {McCullough,Feng,Chen}@ict.usc.edu | Ryan.e.McAlinden.civ@mail.mil |

## ABSTRACT

In the globalized economic world, it has become important to understand the purpose behind infrastructural and construction initiatives occurring within developing regions of the earth. This is critical when the financing for such projects must be coming from external sources, as is occurring throughout major portions of the African continent. When it comes to imagery analysis to research these regions, ground and aerial coverage is either non-existent or not commonly acquired. However, imagery from a large number of commercial, private, and government satellites have produced enormous datasets with global coverage, compiling geospatial resources that can be mined and processed using machine learning algorithms and neural networks. The downside is that a majority of these geospatial data resources are in a state of technical stasis, as it is difficult to quickly parse and determine a plan for request and processing when acquiring satellite image data. A goal of this research is to allow automated monitoring for large-scale infrastructure projects, such as railways, to determine reliable metrics that define and predict the direction construction initiatives could take, allowing for a directed monitoring via narrowed and targeted satellite imagery requests. By utilizing photogrammetric techniques on available satellite data to create 3D Meshes and Digital Surface Models (DSM) we hope to effectively predict transport routes. In understanding the potential directions that large-scale transport infrastructure will take through predictive modeling, it becomes much easier to track, understand, and monitor progress, especially in areas with limited imagery coverage.

## ABOUT THE AUTHORS

**Kyle McCullough** is the Director of Modeling & Simulation at USC-ICT. His research involves geospatial initiatives in support of the Army's One World Terrain project, as well as advanced prototype systems development. His work includes utilizing AI and 3D visualization to increase fidelity and realism in large-scale dynamic simulation environments, and automating typically human-in-the-loop processes for Geo-specific 3D terrain data generation. Kyle received awards from I/ITSEC and the Raindance festival, winning "Best Interactive Narrative VR Experience" in 2018. He has a B.F.A. from New York University. Email: mccullough@ict.usc.edu

**Meida Chen** is currently a research associate at the University of Southern California's Institute for Creative Technologies (USC-ICT) working on One World Terrain project. He received his Ph.D. degree at USC Sonny Astani Department of Civil and Environmental Engineering. His research focuses on the semantic modeling of outdoor scenes for the creation of virtual environments and simulations. Email: mechen@ict.usc.edu

**Andrew Feng** is currently a research scientist at USC-ICT working on the One World Terrain project. Previously, he was a research associate focusing on character animation and automatic 3D avatar generation. His research work involves applying machine learning techniques to solve computer graphics problems such as animation synthesis, mesh skinning, and mesh deformation. He received his Ph.D. and MS degree in computer science from the University of Illinois at Urbana-Champaign. Email: feng@ict.usc.edu

**Ryan McAlinden** is a Senior Technology Advisor for the US Army's Synthetic Training Environment (STE), part of Army Futures Command (AFC). He is the Cross Functional Team (CFT) lead for the One World Terrain (OWT) initiative, which seeks to produce a high-resolution, geo-specific 3D representation of the surface used in the latest rendering engines, simulations and applications. Ryan previously served as Director of Modeling, Simulation &

Training at the University of Southern California's Institute for Creative Technologies where he led several initiatives related to training modernization across the Services. Ryan rejoined ICT in 2013 after an assignment at the NATO Communications & Information Agency (NCIA) in The Hague, Netherlands. There he led the provision of operational analysis support to the International Security Assistance Force (ISAF) Headquarters in Kabul, Afghanistan. Ryan's research interests lie in the design, development, implementation and fielding of solutions that are at the cross-section of 3D rendering, geospatial science, and human-computer interaction. Ryan earned his B.S. from Rutgers University and M.S. in computer science from USC. Email: ryan.e.mcalinden.civ@mail.mil

# Utilizing Satellite Imagery Datasets and Machine Learning Data Models to Evaluate Infrastructure Change in Undeveloped Regions

Kyle McCullough, Andrew Feng, Meida Chen
University of Southern California Institute for Creative Technologies
Playa Vista, California
{McCullough,Feng,Chen}@ict.usc.edu

Ryan McAlinden
Army Futures Command
Orlando, Florida
Ryan.e.McAlinden.civ@mail.mil

## INTRODUCTION

Currently there are a large number of construction initiatives taking place in remote undeveloped regions in the world. Many of these funded by large nations such as Russia, China, and the United States. Some efforts, such as those in the South China Sea to build artificial islands, have a definite military significance, while others seem to lean towards an economic investment strategy, such as China's infrastructure initiatives in Africa, or Russia's $25B loan to Egypt to construct nuclear power plants. While it is difficult to understand direct intention behind any of this investment and construction, understanding how these initiatives have come together and, especially concerning large-scale transport networks, how they are developing is of critical importance to summation of the global geo-political and logistical ecosystem. Advanced and rapid acquisition of knowledge is becoming more and more important as the globalization of numerous industries has now become commonplace. Utilizing satellite imagery to begin to understand some of these initiatives may seem an evident path. However, the sheer magnitude of data in quantity, due largely in part to the size of the globe with the additional temporal scale, means that this analysis becomes incredibly complex. Just a singular imagery resource can reach into the hundreds of petabytes, making selective detection difficult. In this work, we are researching methods for determining the future paths of railways under construction as large-scale transport networks. Predictive modeling of the paths the route may take will reduce the target area for future analysis and greatly reduce the number and size of datasets necessary for analysis. This paper describes our initial photogrammetric reconstruction and artificial intelligence work to analyze the satellite imagery and produce datasets and information that is useful beyond the single case of predictive pathing and to support additional industry and simulation applications.

## BACKGROUND AND RELATED LITERATURE

The first satellite with earth imaging capabilities was the Explorer 6, launched in 1959. Since then a great number of commercial, private, and government satellites with earth imaging capabilities have been placed into earth orbit. Many of these satellites are now able to capture the earth at a very high resolution, less than 50cm Ground Sample Distance (GSD), which is a measure of the space between pixel centers based on ground measurements. There are papers that research many aspects of imagery's application to understanding undeveloped regions, Lower and Middle Income Countries (LMIC), and heavily populated areas. Notably and most nearly related to our work, some researchers have looked in detail at detecting construction starts and tracking development through change detection, for instance through SAR imagery (Jaturapitpornchai et al., 2019 ), and through Very High Resolution satellite imagery, as our research efforts are using (Tian et al., 2016). There are even companies such as Bird.i (S, 2019) that offer this type of work as a service. However, we were unable to find existing research into transport network construction prediction, especially based on data derived from satellite imagery. Though there are a number of transport network predictive models, these take other factors into account, such as economic theory, existing network status, and traffic, as researched in the paper 'Predicting the Construction of New Highway Links' (Levinson & Karamalaputi, 2003), which also relies on possible or existing nodes for making the highway connection predictions. Our effort seeks to use possible nodes coupled with geometry to make long distance determinations, but also to rely on geometry alone for small scales to determine more precise construction paths.

For this work, we will be evaluating DigitalGlobe, specifically the WordView-3 satellite that launched in August of 2014, as a singular provider of satellite imagery for our mesh reconstructions. We will run the imagery through a photogrammetric process in collaboration with Ohio State University (OSU), who have created an optimized pipeline for generating Digital Surface Models and 3D Mesh Geometry from Satellite Data (Qin, 2016). This process will be able to leverage some of our previous research into the satellite imagery reconstruction capability that we have used for efforts on the Army's One World Terrain (OWT) program. (Spicer et al., 2016)

Maxar through DigitalGlobe have amassed over 90PB of data. Typically, this imagery is available for licensing directly; however, we are able to utilize our status as a Department of Defense Sponsored University Affiliated Research Center (UARC) to access the data through the EnhancedView program. EnhancedView is a contractual agreement between the National Geospatial Intelligence agency (NGA) and Maxar (Maxar Technologies, 2019), in which the WorldView-1, WorldView-2, and WorldView-3 satellite data and image archive are made available to groups such as "warfighters, first responders, intelligence analysts and civil government users" to support operations and research (DigitalGlobe, 2013). A number of our efforts into the mapping and simulation capabilities required for geo-specific data already make use of this program, and this access is invaluable to the research we do along a number of project lines, supporting OWT, as well as NGA and the Office of Naval Research.

## SIGNIFICANCE OF THIS WORK

### Tracking Developing Regions

The future battlefield is becoming increasingly complex, with new areas of operation and conflict within space and cyberspace, as well as increased competition to the US and its allies in economic and industrial capabilities. Russia and China are investing heavily in infrastructure in low and middle-income countries (LMICs). Notably many regions in Africa have become a focal point for economic investment, whether through energy, transportation, or natural resource acquisition. It can pay off dividends to begin to understand the purpose and direction of these development initiatives. For instance, using satellite imagery to identify a construction-start can help to model future surveillance of that area, and if that same imagery is used to track a transportation network in-progress, then many predictions can be made about the locations of future development, potential resources, or military encampments. By creating 3D terrain models that can be analyzed in new ways, this type of research enables new dimensions in intelligence gathering and reasoning capabilities. Satellite imagery may reveal a large mining operation in a remote region, but by generating geometric models from that data across a timespan, it is possible to understand the usage, depth, and size of the mine. This type of analytics has not been possible in the past from low cadence 2D imagery alone. Tracking and predicting the construction of large-scale transport networks offers many benefits to understanding the financing, scale, and effort to give a holistic view of infrastructure initiatives.

### Economic & Military Implications

Understanding who and how a group is creating new infrastructure can help to reveal motives in investment and economic overhaul. The idea of a quid-pro-quo when it comes to this type of infrastructure is critical to attempting to reason on the underlying motivations. As an example, Russia loaned Egypt $25 Billion dollars to begin construction on nuclear reactors. While the background intent behind the Russian action is beyond the scope of this paper, the move helped solidify a relationship that resulted in increased Russian military presence and capability within Egypt, including allowing use of airspace and military bases, as well as a number of co-operative agreements (Shaker & Zilberman, 2018). That is a well-documented example, and the knowledge of the construction initiative was well publicized. Had this been a secretive exchange, a system for parsing satellite imagery of the construction could have provided significant intelligence. By utilizing this work to track transport infrastructure our hope is that not only will large-scale construction initiatives such as power plants or resource mines become evident, but that military base locations and equipment transport routes could also be determined before construction has even begun.

### Industrial & Construction Applications

Tracking construction offers a large number of benefits to industry, and the idea of tracking initial constructive initiatives has been covered fully by a number of papers, including an in-depth review by a group of researchers who were able train deep learning networks on satellite acquired synthetic-aperture radar (SAR) imagery (Jaturapitpornchai et al., 2019). Their model detected, with a high level of accuracy, new construction within urban environments using generalized change detection methodology. Our hope with this work is to be able to inform such works to minimize the amount of data required for initial analysis. With relevant prediction based on a limited amount of data, the goal is to narrow required collection timespans and geographic areas based on the predicted path of transport networks. Our work could also help to optimize construction-planning processes. This research seek to find the most efficient route when utilizing navigational mesh geometry, and takes previous construction efforts into account when utilizing machine-learning methods for longer distance predictions. This efficiency model and history-based prediction should allow for a relatively accurate cost estimation based on known construction practices, as well as helping to provide a construction timeline.

**Geospatial Intelligence & Humanitarian**

Attributes of satellite imagery make it very useful 'out-of-the-box' for a number of use cases. Geospatial Intelligence (GEOINT) is one of these use-cases where the benefit is particularly profound. Commercial satellites have a maximum resolution of around 30cm GSD. However, there are highly classified military satellites that may have even smaller GSDs, with optics packages that are capable of resolving down to a few inches. A downside however, is that the time to acquire and size on disk of storage for those resources may inhibit a large-scale analysis. If you know where you want to look ahead of time due to accurate predictive modeling, making a request for high-resolution data becomes not only beneficial, but also more reliable, with fewer resources wasted.

Human Security plays another role within the context of GEOINT. The University of Southern California's Lab for Human Security and Geospatial Intelligence (HSGI) has conducted a number of research initiatives utilizing satellite imagery. One, in partnership with Human Rights Watch, documents destruction of villages in Myanmar via satellite imagery utilizing Artificial Intelligence to analyze the datasets (Marx, Windisch, & Kim, 2019). Their work explores systems and processes close to what we had first begun researching, though with a focus on destruction, as opposed to construction, and high-cadence small satellites as opposed to lower cadence high-resolution satellites.

Being able to predict the paths of transport networks at a large-scale and across vast distances may be applicable to human migrations due to famine, warfare, or other natural disasters. For instance, the HSGI Lab at USC has begun to leverage cellular network data to track large-scale group movements. The research work we are focusing on could assist in determining the predicted path of those migrations, helping to simplify data needs by narrowing the target range, facilitating data acquisition for high cadence satellite imagery, to track the movements through daily visual imagery, and verify the predictions.

**Modeling & Simulation**

There are a number of applications to the modeling and simulation industry, one example is in the area of predictive simulation, using analysis within the simulated environment to make predictions about real world events. Generally, a large amount of data is necessary to increase accuracy in predictive simulation, but in the particular use-case we've chosen to explore for railway construction, our hope is that we can simplify areas for analysis to a very small set of inputs involving the railway itself, and the terrain data. By taking into account general railway construction techniques, such as the length and width of the individual track pieces, along with the specific bounds that define valid terrain on which a rail system can be placed, we hope to use navigation meshes to discern the path of travel between predefined points.

Through this effort, we will also continue to explore photogrammetric 3D Terrain within a simulation environment derived from satellite data. As our 3D Mesh data is generally a much higher visual and spatial fidelity, 2cm GSD in some cases, as generated from low-altitude small-unmanned aircraft systems (sUAS), we hope to understand the strengths and weaknesses of the lower resolution datasets in simulation runtimes. For instance, the satellite derived meshes are especially useful for very large-scale efforts within a simulation environment, and in this work, we are looking at loading around 400sq km of high resolution data into the simulation at any given time for analysis.

**PROCESS**

**Imagery for Photogrammetric Mesh Generation**

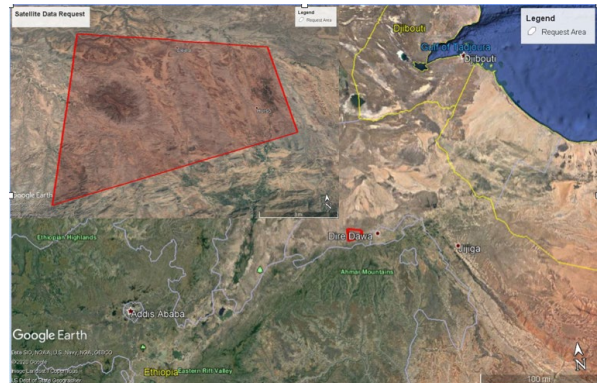

To simplify and verify our intended process for photogrammetric reconstruction and mesh generation, we manually selected an area with a recently constructed railway, where previously very little development existed. Searching through DigitalGlobe imagery using Google Earth, we were able to locate a target area, by using the image display and temporal capabilities of the Google Earth Pro desktop tool. We decided to choose a relatively small region with inherent geometric features and elevation changes that would directly affect the construction path of a transit line. Ethiopia has recently undergone substantial development due to investment from the Chinese government and the effect on the region is evident from satellite imagery even at the lowest available resolutions. By tracing the newly constructed

**Figure 1. Extents Determined within Google Earth for Satellite data Request.**

Addis Ababa-Djibouti Railway line, we were able to find an area near Hurso, Ethiopia to use as a relevant test case for our work, see Figure 1. By limiting the scope of our initial efforts to railway lines and existing construction, we were able to gather relevant data and test our methods with the ability to verify against the ground truth of the actual railroad construction.

**Imagery Collection & Analysis**

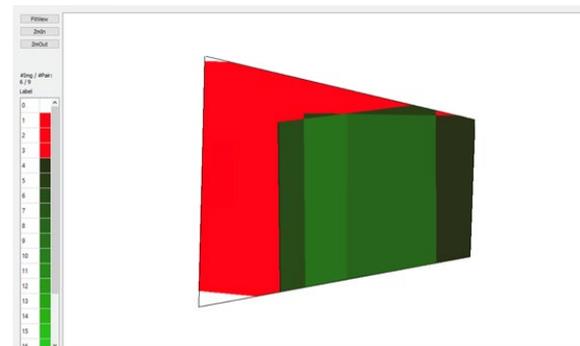

After determining the target location, we created a Keyhole Markup Language (KML) file to create the bounds of the dataset and began working with our collaborators at Ohio State University (OSU) to determine that the relevant data was available through EnhancedView. EnhancedView offers a simple web-based front end for the Maxar library that allows searching by image ID and then downloading the raw satellite imagery in a number of formats. The total imagery required for a complete photogrammetric reconstruction and initial analysis was over 500GB. Once we'd acquired the imagery, OSU was able to run it through their analysis tools, to determine the quality of the stereo pairs, a process that is detailed in their work, "A critical analysis of satellite stereo pairs for digital surface model

**Figure 2. Visual Analysis of sufficient quality and stereo pairs for source imagery.**

generation and a matching quality prediction model" (Qin, 2019) and the analysis result is visualized in Figure 2. OSU then identified the sufficient data, and processed the relevant imagery through their RPC (rational polynomial coefficient) stereo processor (RSP) software pipeline, which creates DSM data and Geometry based on RPC modelled satellite imagery. Their process includes geo-referencing, point cloud generation, pan-sharpening, DSM resampling and ortho-rectification (Qin, 2016) and has been enhanced to support fully automated processes for 3D Recovery. (Qin, 2017)

One major problem that became immediately apparent was that within the area we had chosen for proofing, existing coverage from Maxar's WorldView cluster of satellites was lacking within the periods we had hoped to analyze. One major reason for selecting this area was the particular geometry, but also the year in which the railroad was constructed through this region was recent enough to hopefully have sufficient satellite coverage in which we could explore the reconstructed data in three periods, before, during, and after the railroad construction. In fact, there was decent coverage of the area itself, but it was difficult to find matching pairs from various imaging fly-overs that was useful for the reconstruction (Figure 3). There were multiple views of pre-construction and post-construction, however there was limited data available during the initial construction phase, which meant we could potentially only generate a few temporal variations. This limitation was not difficult to overcome for the scope of our research work; however, it is important to keep in mind for future efforts applying this research in regions that are not photographed regularly with high-resolution satellites.

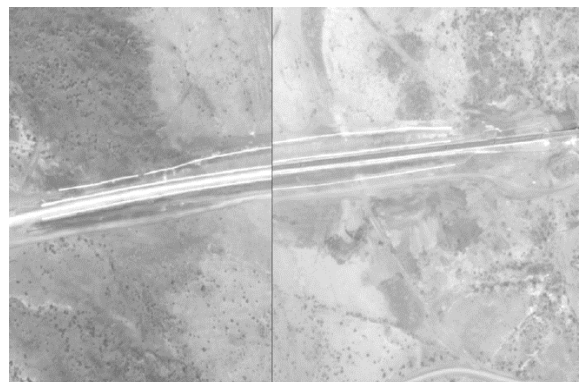
**Figure 3. Unmatched Pairs based on available satellite data**

**Imagery Processing**

There is a large amount of research and commercial applications surrounding the process of photogrammetric reconstruction. Our team has typically focused on reconstruction of geo-specific terrain via drone collection (Spicer et al., 2016). Our research goals toward source-agnostic data for the One World Terrain pipeline have led us to work with a number of additional groups that have been utilizing photogrammetry techniques for satellite data, notably Vricon and OSU, exploring data fusion of the various types.

As previously described, OSU's satellite imagery reconstruction pipeline allows for all of the outputs we need for this research, covered in two common terrain data formats. One is the Digital Surface Model (Figure 4), which we will use as a GEOTIFF, and the other is a 3D Mesh Model (Figure 5) which is generated in the OBJ format, and then reprocessed to a more runtime optimized mesh format. This optimized format is based on a binarized form of in-memory mesh within the Unity Game Engine. This binarized form is Unity specific and called a Lockheed Martin Asset Bundle (LMAB). Once we had the sample area as both a textured mesh model and a DSM, we were able to begin testing our two approaches to see if we could automate the analysis required to predict a transport network path.

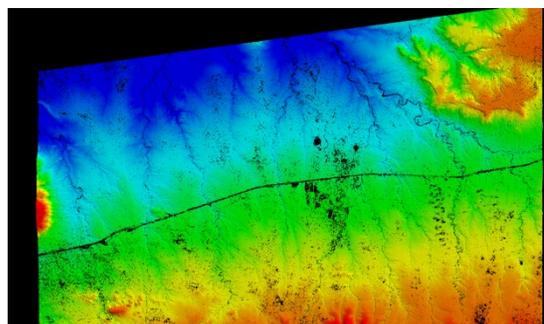
**Figure 4. Colorized DSM from Satellite Reconstruction**

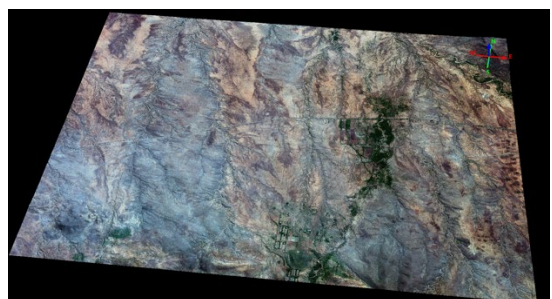
**Figure 5. Mesh Reconstruction from Satellite Imagery**

**Digital Surface Models for Machine Learning Methods**

The second part of this effort was to look at utilizing Digital Surface Models (DSM) within a machine learning process that could leverage prior work we have done to predict smaller scale pedestrian movement. For this, we needed a large amount of DSM data. Though we could use the reconstruction pipeline at OSU to derive Very High Resolution (VHR) DSM data, the time it takes to generate the VHR DSMs and the amount of satellite imagery required, was not optimal to be able to provide as data for our Machine Learning data models. Machine learning generally requires an enormous amount of data to be effective, and that data doesn't necessarily need the resolution offered by the RSP process. We determined a reasonable level for the resolution needs and were able to utilize NASA's Shuttle Radar

Topography Mission (SRTM) data available through NGA's Geospatial Repository & Data Management System (GRiD) which offers an easy-to-use intuitive interface for accessing geospatial data. The SRTM data we were requesting covers the entire planet, and the lower resolution DSM means that we are able to download extremely large areas quickly, in a very efficient size on disk. We were able to utilize an area covering the entire range of the Djibouti-Addis Ababa Railway in under 50MB. This data was downloaded from GRiD directly, in an imagery format ready for ingestion into our machine-learning pipeline, and required no further processing before being utilized by our machine learning models.

**CURRENT WORK**

Our work attempts to utilize two separate methods to understand the viability of route prediction based on terrain properties that using only the shape and relief of the datasets. Both methods utilize aspects of the terrain such as elevation, slope and general topography. Our goal is to utilize the machine learning models for long-range prediction, and couple that with a navigational mesh (NavMesh) prediction for shorter and more precise geometric pathing. As defined by Unity, a NavMesh is a data structure, which describes the walkable surfaces of a game world and allows finding a path from one walkable location to another (Unity Technologies, 2020). We theorized that we could use the methodology with some base parameters to replicate the path of a railway as well. Our future work seeks to prove that a combination of both methods would provide the most accurate predictions for construction positions and direction.

**Path Prediction via Navigational Meshes**

For our first experimental method, we decided an accurate prediction might be attained by borrowing a pathfinding solution from the commercial game industry, utilizing our prior experience on navigational meshes within a runtime environment (McCullough et al., 2019). In the case of both the previous and current work we used the A* Pathfinding algorithm and Unity 3rd Party Library by Aron Granberg (Granberg, 2017) to create a recast graph within the Unity game engine. In our previous work, we attempted to use a NavMesh to guide a large number of autonomous agents over both complex urban terrain as well as large undeveloped regions. Our hope with our current work was that by utilizing factors that would be most relevant to railroad construction we could understand within some degree of certainty how to define the parameters for the A* algorithm to suit our construction use-case. While a number of railroad construction methods and techniques could vary between companies as well as by construction year, by focusing solely on two key factors we hoped to be able to mimic a traversable path. These factors are the approximate width of railroad tracks and the maximum grade a rail system can traverse and remain effective. We determined that approximating these figures would be enough based on the small grade variation in railroad design – for instance, the maximum slope for freight trains should generally be under 1% for a main line, and grades steeper than 2.2% are rare (McGonigal, 2006) which makes for a restrictive parameter.

The first step was to take the Tiled OBJ generated from the Satellite data and ingest it into Unity. Once we'd ingested the Tiled OBJs into a workable terrain set within our runtime environment, we were able to create the NavMesh, in this case we defined the parameters and used A* graph methodology to define the maximum traversable slope and the width of the railway.

Once the NavMesh was created, we were able to utilize the Aerial Terrain Line of Sight Analysis System (ATLAS) tools we had previously designed for route planning and line of sight awareness (USC ICT, 2017). If we knew the location of a nearby major city, along with the tendency for a train to maintain a straight path for highest speed travel, we could begin to find a number of target points the railway may ultimately take. Our goal was not to find the route with an extremely accurate spatial fidelity, but rather to determine the probability that the construction would take a particular route to inform where additional data could be pulled from. As a single region of around 100sqkm can contain hundreds of gigabytes of high-resolution imagery that would require parsing, it becomes much more effective to target smaller areas where the construction is expected to take place to follow the route temporally via satellite imagery. To test the path prediction we started with a ground truth point where the railroad was traveling straight and continued that line directly across a 10km area to choose the ending point for the pathfinding tool. Using this method definitely proved useful in utilizing the photogrammetric geometry to create a NavMesh and simulate a prediction path. As seen in Figure 6, the result was able to take into account some of the finer geometric details and route around them. We are confident that with additional optimization of the pathfinding parameters and future work to take into account construction methodologies and semantic classification we could improve the result further using solely the photogrammetric mesh method.

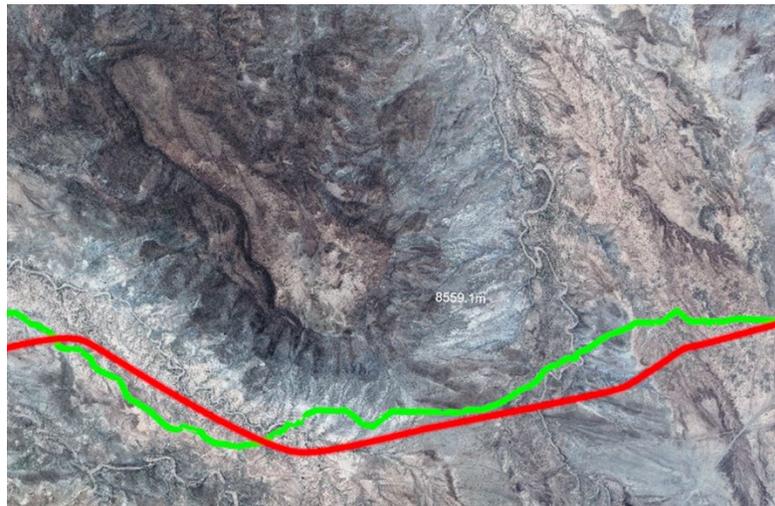

**Figure 6. The green line represents the NavMesh path prediction. The red line represents the railway ground truth.**

**Path Prediction Via Machine Learning**

As in other prediction tasks, machine-learning methods are well suited to future path prediction. Previous research related to this work focused on predicting a person's future trajectory given recent observations. (Feng & Gordon, 2019) There are two main differences and challenges when applying this prior art for our proposed application of railway path predictions. First, unlike a pedestrian path, the railway path is larger in scale and tends to cover hundreds of miles of area for possible future path predictions. Moreover, using only the previously observed path as input ignores the factors contributed by surrounding environments, which plays an important part when deciding the feasible railway path. Therefore, it requires an efficient way of integrating terrain information into the learning process to best predict the future construction path.

Modern neural networks offer several approaches such as variational auto-encoder (VAE) to extract compact features from image data without supervision. With enough training data and computational resources, such an architecture could learn generalized features of the terrain that are most informative and help the integration of such terrain information into the path prediction process. A relevant previous work (Feng & Gordon, 2019) describes a method of encoding terrain digital elevation models (DEM) as vectors in latent space, and their use for predicting future trajectory positions. Their method is separated into two stages. In the first stage, Wasserstein Autoencoder is applied to obtain the latent space from nearby terrain digital elevation models (DEM). Then the latent variables obtained from the DEM patches are concatenated with the corresponding past trajectory coordinates to form the input vector for the next stage. For the second stage, a convolutional neural network is used to predict the future trajectory positions from past trajectory and corresponding terrain latent space. Their experiments show accuracy gains with the inclusion of latent terrain representations in trajectory prediction. Therefore, a feasible approach using machine learning is to apply the aforementioned framework to train a path prediction model for railway constructions, taking account of nearby terrain information through both satellite imagery and DEM.

In our experiment, we formulate the railway prediction task as an Deep Inverse Reinforcement Learning (IRL) problem over the DEM features, which has been successfully used to predict traversability cost maps for path planning in urban settings (Wulfmeier et al., 2016) or inverse control of robots (Finn et al., 2016). Specifically, we apply IRL to learn a cost map for predicting new railway trajectories over a DEM terrain based on demonstrations of past railway constructions in the area. Overall, the method iterates the following two steps until convergence: the cost estimation step to recover the underlying cost function map, given demonstrations; the policy estimation step uses the current estimate of the cost map to solve the forward path prediction. The cost estimation step first uses a convolutional encoder-decoder network such as U-Net (Ronneberger et al., 2015) to transform the DEM terrain into a traversability cost map. In the policy estimation step, the cost map generated from network is used to predict a railway path based on A* algorithm. Finally, the error between the predicted path and ground-truth is backward propagated through the U-Net to refine the cost map estimation. Figure 7 shows the resulting cost map and predicted railway from this process. The results show that the IRL process is able to extract interesting terrain geometric features to produce the cost map, and the predicted railway is reasonable compared to the ground truth. Since the method performs on 2D DEM, it is more scalable for prediction over a long trajectory in large areas. It also does not require 3D reconstructions to build the navigation mesh, and therefore is suitable for cases when multi-view satellite imagery is not available for the region of interest or acquiring a very large amount of data with time to process, it is prohibitive.

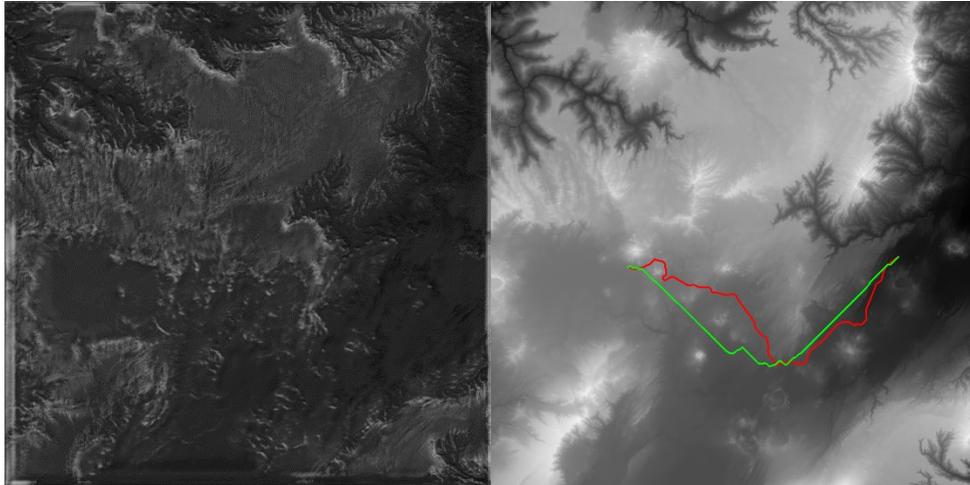

**Figure 7. (Left) The estimated cost map using U-Net. (Right) Green line represents the IRL path prediction. Red line represents the railway ground truth.**

**FUTURE WORK**

We were not able to generate models of sufficient quality for each of the temporal phases we'd hope to understand, covering the time before, during, and post-construction. For our future work, we expect to spend more time investigating control areas in which the flyover is sufficient during the construction period to generate photogrammetric reconstruction of each temporal phase. We also believe that this problem could be solved by using the higher-cadence smallsats for photogrammetric reconstruction, though this might present new challenges as the imagery is lower resolution. However, for the scope of our current work, we were able to generalize the terrain geometry and utilize the existing imagery to understand our results. In addition, we were able to discern additional ground truth from additional sources, such as publicly available vector data for the ultimate path of the railroad and were easily able to verify our processes.

Though we have methodology for georeferencing the terrain, it was not used in this particular research, as we did not utilize factors relevant to the geospatial data. However, there is potential for tying the spatial reference information to aspects of the temporal simulation, for instance, using the global location to pull weather information that could be used to inform construction speed and delays. In addition, with inherent global geospatial positions, we could more accurately evaluate the results against true accuracy.

Additional work could also use the geographic location along with a material classification pipeline similar to the one we've used in our previous work (Chen et al., 2019) to create enhanced models of the A* pathfinding navigation graph, that would be able to determine the difficulty of leveling terrain, or the solidity of the surface material being constructed upon. There would be a large difference in the construction time and procedure when it came to routing a new railway across soft sand opposed to solid rock. We believe that by taking a deeper dive into construction practices and methods, as well as considering these materials in the predictive model, we could greatly improve the accuracy of our predictions. However, for this work, we focused solely on the terrain relief through the reconstructed geometry alone.

**CONCLUSION**

This work achieved a positive result for general prediction, and considering our limited parameters for the NavMesh method, and our limited training data and DSM resolution for our machine-learning methods, increasing the accuracy of the prediction should only be a matter of providing improved inputs. The foundational techniques appear to be sound. Photogrammetric reconstruction via satellite imagery for NavMesh generation provided interesting short distance results considering the process only used the geometry and simplified navigation and construction parameters. The machine learning method also provided a reasonable result considering we only used a relatively small sample of 2D DSM data at 60m resolution to train the model. We believe that by improving resolution, utilizing additional training data, considering material classification, and combining the two methods we could greatly increase the accuracy and value of the prediction.

**ACKNOWLEDGEMENTS**

The authors would like to thank the two primary sponsors of this research: Army Futures Command (AFC) Synthetic Training Environment (STE), and the Office of Naval Research (ONR). This work is supported by University Affiliated Research Center (UARC) award W911NF-14-D-0005. Statements and opinions expressed and content included do not necessarily reflect the position or the policy of the Government, and no official endorsement should be inferred.

The authors would also like to graciously thank Rongjun Qin and his team at Ohio State University for providing us with Satellite Reconstructions in Mesh and DSM Formats for this work, as well as Andrew Marx and Steve Fleming at SSI, for their assistance and help in sourcing data, providing advice, and discovering a unique research goal.